\documentclass[letterpaper, 10 pt, conference]{ieeeconf}  

\IEEEoverridecommandlockouts                              

\usepackage{cite}
\usepackage{amsmath,amssymb,amsfonts}
\usepackage{algorithmic}
\usepackage{graphicx}
\usepackage{textcomp}
\usepackage{xcolor}
\usepackage{multirow}
\usepackage{bm}
\usepackage{float}

\usepackage{hyperref}
\usepackage{xurl}

\title{\LARGE \bf
A Black-Box Positioning Solution for Effortless Integration in \\ Wearable Navigation}

\author{Anh-Van Vu, Changmin Sung, Dongsoo Han, Member, IEEE
\thanks{This research was supported by the Challengeable Future Defense Technology Research and Development Program through the Agency For Defense Development(ADD) funded by the Defense Acquisition Program Administration(DAPA) in 2025(No.915027201) and the Technology Innovation Program (RS-2025-02313957) funded By the Ministry of Trade, Industry \& Energy(MOTIE, Korea).}%
}

\begin{document}

\maketitle
\thispagestyle{empty}
\pagestyle{empty}


\maketitle
\begin{abstract}

Accurate localization is critical for a wide range of applications, including navigation, augmented reality, safety management, and healthcare. However, achieving accurate positioning remains challenging, especially in GPS-degraded environments. Pedestrian Dead Reckoning (PDR), which estimates position based on inertial sensor data, offers a promising alternative but typically demands substantial effort in algorithm design and system integration. This complexity often leads to prolonged development timelines and elevated costs. To address these challenges, we present the PDR Sensor (PDRS), a plug-and-play, black-box PDR module optimized for seamless integration into wearable and IoT platforms. PDRS integrates a microprocessor and an inertial measurement unit (IMU) within a compact, low-power module. It executes real-time sensor fusion onboard, providing position updates via standard interfaces (I²C, SPI, UART). Experimental evaluations demonstrate its high robustness, achieving a step detection error of just 0.16\% over 1250 steps and a traveled distance error of 1.87\% under diverse conditions. The system further exhibits low latency (2.5 ms) and modest power consumption (52.35 mA). Unlike conventional studies, the PDRS leverages a holistic co-design of hardware and software to enhance performance and integration aspects, simplifying adoption without requiring deep expertise. By lowering technical barriers, PDRS enables developers and researchers to prioritize innovation in IoT and wearable technologies, accelerating prototyping, reducing development time, and advancing industry progress. Future work aims to enhance accuracy, reduce power usage and module size, and integrate AI-driven techniques to better mitigate drift.

\end{abstract}

\section{Introduction}

Indoor positioning, crucial for diverse applications ranging from navigation in complex environments (e.g., airports, shopping malls) to augmented reality and healthcare \cite{han2014building, kamiya2019indoor, hu2022experience}, relies heavily on PDR. Its infrastructure-free nature, utilizing only inertial sensors (accelerometers and gyroscopes) to track user movement, offers a significant advantage where satellite or RF methods are ineffective \cite{zhou2023gps, ojeda2007personal}. This autonomy, achievable with compact IMUs integrated into modern smartphones and wearables \cite{harle2013survey, hou2020pedestrian}, has fueled extensive research resulting in numerous PDR algorithms and system designs \cite{wu2019survey, harle2013survey, hou2020pedestrian, 6987239, jimenez2009comparison}. Therefore, it is now an essential component of most indoor localization systems\cite{guo2020improved,zhou2023gps, chen2015integrated, kong2022hybrid, el2021indoor}. 

Despite significant advancements, deploying practical PDR systems remains challenging. PDR is a relative positioning method prone to cumulative drift errors over time \cite{tong2019robust, wu2019survey, harle2013survey, potorti2020ipin}, where even minor sensor biases or step length inaccuracies compound, causing trajectory divergence. Consequently, standalone PDR is unsuitable for long-term tracking, necessitating fusion with external references or additional sensors (e.g., magnetometers, barometers, UWB, map constraints) to mitigate drift \cite{potorti2020ipin, guo2020improved,zhou2023gps, chen2015integrated, kong2022hybrid, el2021indoor}. However, such multi-sensor fusion increases system complexity, power consumption, and cost, undermining PDR's inherent simplicity. Practical indoor positioning demands ease of implementation, affordability, and minimal computational complexity \cite{tomavzivc2021indoor}, criteria often unmet by research prototypes requiring specialized hardware or extensive calibration.

Another key challenge involves the significant expertise and development effort required to create robust PDR solutions. Building a PDR system necessitates real-time coordination of numerous components on resource-constrained hardware: sensor data acquisition, signal processing (step detection, orientation and distance estimation), and error correction. As Nilsson et al. observe, although ``dead reckoning is not rocket science," the number of hardware/software components that must work in harmony is large, and achieving a well-performing embedded implementation is far from trivial \cite{nilsson2012foot}. Careful selection of an embedded processor with sufficient processing power and algorithmic optimization for low latency and energy consumption is crucial.

Further complexities arise from the need for small form factors and mechanical reliability (for wearable or mobile applications), with few readily available modular, reusable hardware-algorithm solutions \cite{nilsson2012foot}. Commercial solutions, such as Bosch Sensortec’s BHI160B and BHI260AP, can fulfill these requirements but require specialized PDR-enabled firmware, which is often publicly unavailable \cite{BHY2_SensorAPI}. This necessitates significant custom integration, increasing development time and costs for solutions seeking to utilize PDR. Therefore, despite the ubiquity of low-cost MEMS sensors, the barrier to entry for deploying PDR systems remains high. This gap between algorithmic research and deployable products, highlighted in recent literature \cite{tomavzivc2021indoor}, emphasizes that accuracy alone is insufficient; stability, real-time capability, energy efficiency, and easy integration are equally critical for widespread adoption.

This paper addresses the challenges of deploying PDR systems by introducing the PDR Sensor (PDRS), a compact, low-power hardware module integrating a microcontroller unit (MCU), an inertial measurement unit (IMU), and an onboard adapted PDR algorithm. Designed as a plug-and-play solution, the PDRS encapsulates complex sensor fusion and computation logic, providing real-time position updates via standard interfaces (I²C, SPI, UART). Our module can be treated as an off-the-shelf component – effectively a “black box” PDR unit – that researchers, and system designers can drop into their prototypes or products with minimal effort. This hardware-software co-design contrasts with previous research primarily focused on algorithmic accuracy within high-end processing environments \cite{6987239, wu2019survey, harle2013survey, potorti2020ipin}. By offloading computational complexity and sensor management, the PDRS enables developers to obtain accurate step counts, heading, and displacement estimates without specialized signal processing or inertial navigation expertise. Utilizing inexpensive, commercially available, reproducible components, the PDRS achieves competitive accuracy (0.16\% step detection error, 1.87\% travel distance estimation error) with low latency (milliseconds) and power consumption (tens of milliamps), suitable for numerous real-time positioning applications. This simplified PDR integration significantly lowers the barrier to entry for indoor localization, accelerating adoption in applications ranging from IoT devices to assistive wearables, where a self-contained inertial positioning module unlocks new functionality without added cost or complexity.

The remainder of this paper is organized as follows: Section \ref{hardware and software} details the hardware and software co-design of the PDRS. Section \ref{pdr} describes the adapted PDR algorithm and bias mitigation technique. Section \ref{experiments_and_results} presents evaluations of positioning accuracy in different environments, real-time latency, power consumption, and thermal characteristics. Finally, Section \ref{conclusion_and_discussion} concludes and outlines future works.

\section{Related Works}
PDR has long been studied as a viable solution for indoor positioning when external signals are unavailable. Early work by Beauregard and Haas \cite{beauregard2006pedestrian} established the basic framework of PDR—leveraging step detection, stride estimation, and heading determination from wearable sensors. Their approach, however, required careful calibration to accommodate individual gait variations and stable sensor mounting, underscoring intrinsic limitations that would drive subsequent research.


A key challenge in PDR is drift caused by sensor biases and noise. Ojeda and Borenstein \cite{ojeda2007personal} addressed this using zero-velocity updates (ZUPT), exploiting the foot’s stationary phase to reset errors and improve short-range tracking accuracy. Nilsson et al. \cite{nilsson2012foot} further advanced this concept with OpenShoe, embedding an IMU and microcontroller in footwear for continuous ZUPT-aided inertial navigation. Despite effectiveness, these foot-mounted systems remained sensitive to sensor placement, requiring robust stationary-phase detection, prompting refinements through probabilistic filtering methods like Extended Kalman Filters (EKF) and heuristic drift-reduction techniques \cite{hou2020pedestrian, wu2019survey}. Moreover, the reliance on precise placement and individual calibration highlighted the gap between research prototypes and practical, user-friendly solutions.


Beyond foot-mounted systems, smartphone-based and wearable PDR solutions have expanded to integrate a broader sensor array \cite{6987239, hou2020pedestrian}. Modern approaches typically fuse accelerometer, gyroscope, and magnetometer data to estimate displacement and heading. Hou and Bergmann \cite{hou2020pedestrian} reviewed various wearable PDR systems, noting that although sensor fusion enhances accuracy, indoor magnetic disturbances and variable sensor orientations often require frequent recalibration. Many systems mitigate these issues by periodically incorporating external references or adaptive filtering, yet these methods increase complexity and computational overhead.

Recent advancements have explored data-driven and AI-based approaches to further enhance PDR performance \cite{saha2022tinyodom, sun2021idol}. Saha et al. \cite{saha2022tinyodom} introduced TinyOdom, a framework that employs temporal convolutional networks (TCNs) to process raw IMU data and output displacement estimates in real time. By integrating learned motion patterns with physics-based constraints, TinyOdom achieved competitive accuracy on resource-constrained devices, demonstrating that neural models can complement traditional signal processing. However, the necessity for extensive training data and computational complexity remains a challenge for widespread deployment in consumer devices.


Industrial developments further underscore the trend toward integrated, user-friendly PDR solutions. Companies like STMicroelectronics have developed turnkey PDR engines optimized for smartphones, providing developers with Android and iOS libraries that fuse data from ST’s MEMS sensors to automatically adjust for sensor error, offering PDR as well as contextual awareness \cite{ST_PDR_Blog_2016}. Similarly, Bosch Sensortec’s BMI270 \cite{bosch_bmi270} and ST's LSM6DS3 \cite{st_lsm6ds3trc_ds} incorporate on-chip algorithms for step counting and activity recognition, offering low-power, plug-and-play functionality that simplifies integration for wearables. Although these sensors alleviate some computational burdens, they typically deliver only partial functionality (e.g., step detection without full trajectory estimation). More recently, several smart sensors from Bosch Sensortec, including BHI160B and BHI260AP \cite{bosch_bhi260ap}, have been promoted as fully PDR-capable. However, these capabilities require dedicated PDR-enabled firmware, which remains unavailable \cite{BHY2_SensorAPI}.


In summary, prior research has made significant strides in reducing drift, integrating sensor fusion, and applying AI to enhance PDR capabilities. Foot-mounted systems offer high accuracy with ZUPT but demand careful calibration, while smartphone-based solutions use multi-sensor fusion at the cost of increased complexity. Recent industry efforts have shifted toward integrated sensor solutions, though these often address only parts of the broader PDR problem. Our work tackles these gaps with a compact, black-box PDRS that encapsulates PDR complexities within a low-power module, enabling seamless integration into wearables and IoT devices without requiring specialized expertise. By aligning and extending recent advances, the PDRS provides a robust, accessible solution bridging high-performance academic prototypes and commercially viable systems.

\section{Hardware and Software Co-Design} \label{hardware and software}

\subsection{PDRS Stack} 

\begin{figure}[t]
  \centering
  \includegraphics[width=80mm]{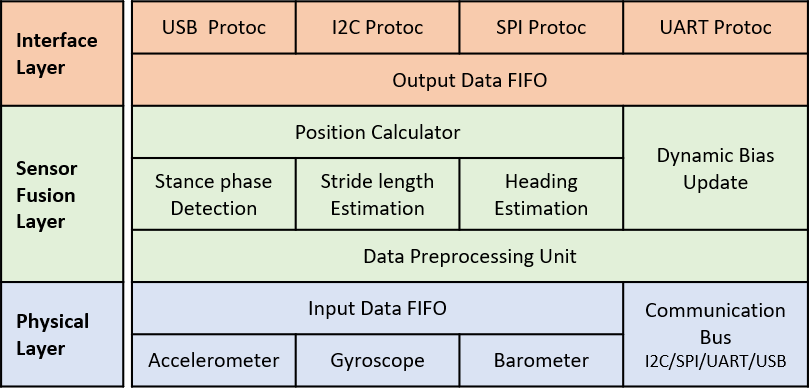}
  \caption{PDR sensor stack}
  \label{fig:pdr_sensor_stack}
\end{figure}

The stack (Figure \ref{fig:pdr_sensor_stack}) embodies the software-hardware co-design architecture of the PDRS, structured into three integrated layers. The Physical Layer contains an accelerometer (measuring linear acceleration), a gyroscope (tracking angular velocity), and a barometer (detecting elevation changes) with a FIFO buffer to manage high-frequency sensor data collection, alongside physical buses for connecting with other systems. The core Sensor Fusion Engine Layer processes raw data through preprocessing (noise filtering, axis alignment, calibration), gait phase detection (identifying stance moments), stride length estimation (adapting to speed/terrain variables), heading calculation (fusing gyro/accelerometer data), trajectory synthesis, and dynamic gyroscope bias update to mitigate drift, ensuring accurate and reliable positioning over time. The Interface Control Layer standardizes communication via I²C/SPI/UART/USB protocols, provides an output FIFO memory for real-time location data access, and enables raw sensor data retrieval, ensuring plug-and-play compatibility across a wide range of devices and systems. This stack transforms multi-sensor inputs into actionable positioning through coordinated hardware resource management and algorithmic computation, balancing accuracy with modular integration and ease of deployment.

\subsection{Hardware Design}

\begin{figure}[b]
  \centering
  \includegraphics[width=80mm]{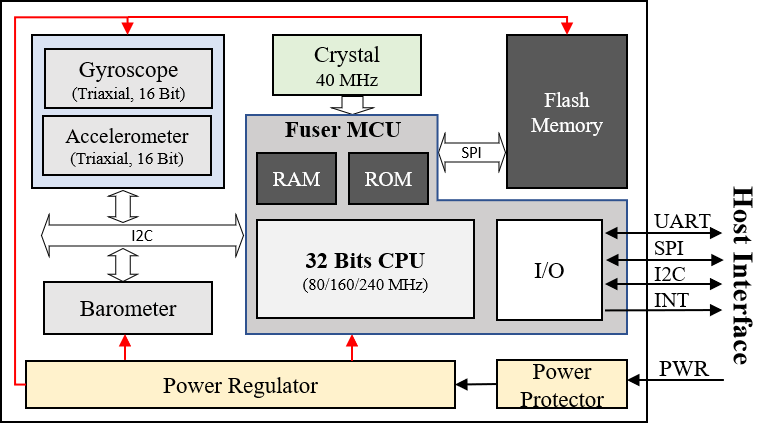}
  \caption{PDRS hardware block diagram}
  \label{fig:sensor_hardware_block_diagram}
\end{figure}

\begin{figure}[t]
  \centering
  \includegraphics[width=60mm]{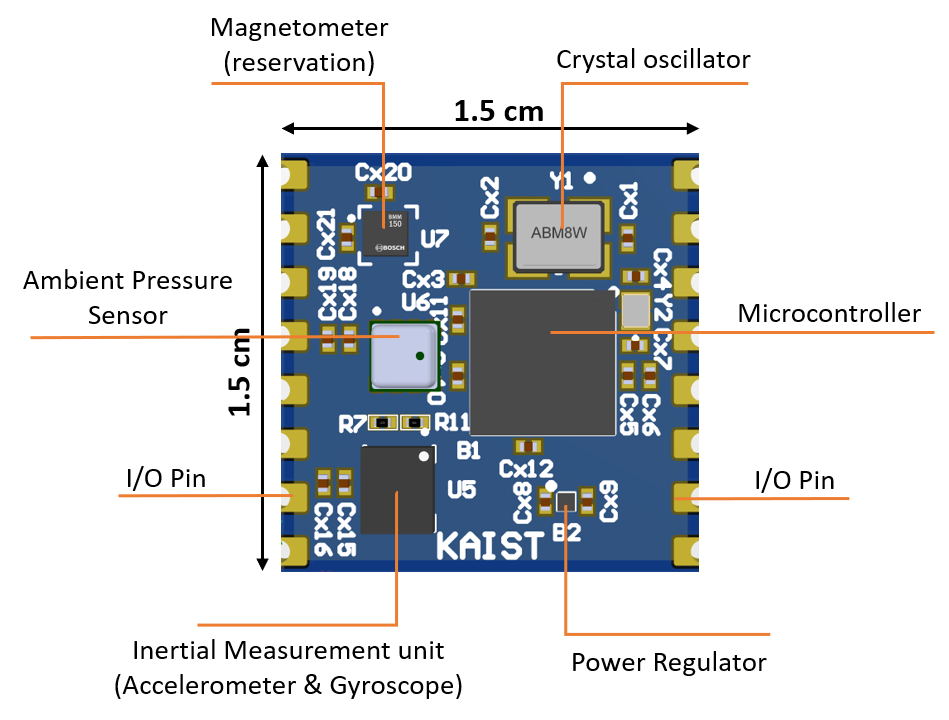}
  \caption{PDR Sensor hardware}
  \label{fig:vpdr_sensor}
\end{figure}

To implement the PDRS stack, a dedicated compact hardware module was designed, incorporating essential components and sensors. The block diagram of the hardware is illustrated in Figure. \ref{fig:sensor_hardware_block_diagram}. At its core, a 32-bit MCU serves as the primary computational element, coordinating peripheral operations, executing sensor fusion algorithms, and managing external communication. The MCU's clock source is a precision crystal oscillator. The sensing subsystem comprises an accelerometer, a gyroscope, and a barometer, which collectively provide data for location estimation. These sensors interface with the MCU via a shared bus. For program storage and configuration parameters, flash memory is integrated, connected to the MCU through a high-speed quad-SPI interface. Additionally, the PDRS module provides standard serial communication buses, including UART, SPI, and I2C, to enable interoperability with external master systems. Dedicated pins are also allocated for auxiliary functions such as data-ready interrupts, status indication, and clock synchronization. Power management is achieved using an ultra-low dropout (LDO) regulator paired with a protection circuit to ensure resilience against short circuits, over-voltage conditions, and other electrical anomalies. The use of LDO regulator minimizes energy loss while maintaining stable sensor operation. The practical design of the PDRS module is depicted in Figure. \ref{fig:vpdr_sensor}, while Table \ref{tab:hardware_description} summarizes the key components and their specifications.

\begin{table}[t]
\renewcommand{\arraystretch}{1.3}
\caption{Hardware specification of PDRS}
\label{tab:hardware_description}
\resizebox{\columnwidth}{!}{%
\begin{tabular}{|c|l|l|l|}
\hline
\textbf{\#} & \multicolumn{1}{c|}{\textbf{Components}} & \multicolumn{1}{c|}{\textbf{\begin{tabular}[c]{@{}c@{}}Manufacturer Product \\ Number\end{tabular}}} & \multicolumn{1}{c|}{\textbf{Features}}                                                                                                                                \\ \hline
1           & MCU                                      & ESP32                                                                                             & 32 bits Dual-core XTensa LX7                                                                                                                                         \\ \hline
2           & Accelerometer                            & \multirow{2}{*}{BMI270}                                                                              & \begin{tabular}[c]{@{}l@{}}16 bit digital, triaxial \\ range: ±2g/±4g/±8g/±16g\end{tabular}                                                                          \\ \cline{1-2} \cline{4-4} 
3           & Gyroscope                                &                                                                                                      & \begin{tabular}[c]{@{}l@{}}16 bit digital, triaxial \\ range: ±125/ ±250/ ±500/ ±1000 /±2000dps\end{tabular}                                                         \\ \hline
4           & Barometer                                & BME280                                                                                               & \begin{tabular}[c]{@{}l@{}}Pressure: 300 to 1100 hPa (Res 0.18 Pa)\\ Temperature: -40 to 85 °C (Res 0.01 °C)\\ Humidity: 0 to 100 \%RH (Res 0.008 \%RH)\end{tabular} \\ \hline
5           & Flash Memory                             & W25Q01JVZEIQ                                                                                         &  Max data transfer rate: 66MB/S                                                                                                                              \\ \hline
6           & Oscillator                               & \begin{tabular}[c]{@{}l@{}}ABM10W-40.0000MHZ\end{tabular}                                & 40 MHz ±10ppm                                                                                                                                                        \\ \hline
\end{tabular}%
}
\end{table}

\subsection{Firmware Design}

\begin{figure}[t]
  \centering
  \includegraphics[width=85mm]{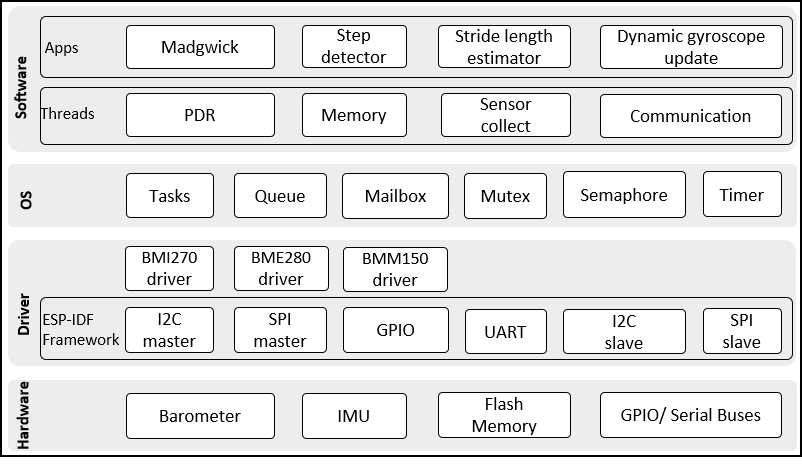}
  \caption{RTOS-based firmware structure of the PDRS}
  \label{fig:RTOS-based-firmware-structure}
\end{figure}

\begin{figure}[b]
  \centering
  \includegraphics[width=85mm]{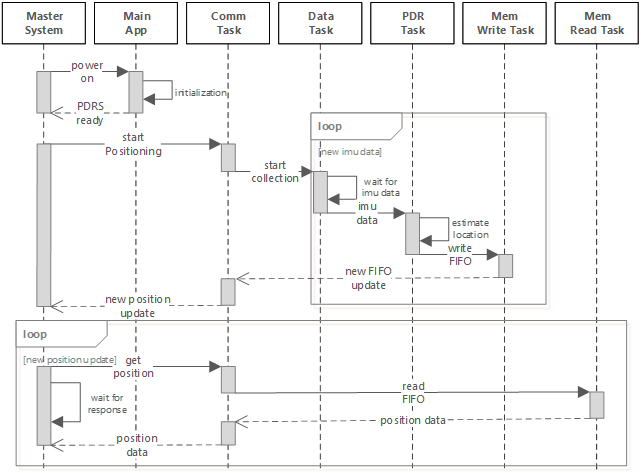}
  \caption{Sequence diagram of the RTOS-based PDRS's firmware}
  \label{fig:task_interaction}
\end{figure}

The PDRS firmware is built on FreeRTOS and organized as a hierarchical stack to maximize modularity and efficiency on resource‑constrained hardware. As shown in Figure.\ref{fig:RTOS-based-firmware-structure}, the Hardware layer contains on‑board sensors and peripherals, while the Driver layer abstracts their device‑specific details. Above them, the OS layer employs FreeRTOS primitives for deterministic scheduling and synchronization. At the apex, a unified Software layer blends system threads with application logic to manage data flow, memory, communication, and the core positioning algorithms. In practice, each functional thread is implemented as an individual OS task. These tasks operate with distinct priority levels and are capable of inter-task communication. The interaction between tasks as well as between the PDRS and the host system is illustrated in the sequence diagram in Figure \ref{fig:task_interaction}, highlighting the coordination mechanism and system-level integration.

\subsubsection{Data Acquisition Task}

interacts with the physical sensors to manage the data collection. The quality of the data directly impacts the accuracy of the positioning result. Consequently, the data must be both accurate and precisely time-stamped. To meet these stringent requirements, we implement an interrupt-driven approach that tightly couples the data acquisition task with an interrupt handler triggered by data-ready signals of the IMU sensor. This interrupt handler has highest priority, ensuring immediate responses when the sensor data is available. This coordination enables minimal latency and packet loss during data collection 

\subsubsection{PDR Task} is responsible for processing sensing data received from the data acquisition task and storing the estimated location (or raw sensing data if needed) in the FIFO memory. The application's host controller can access this data via a serial communication channel managed by the communication task. The cornerstone of the PDR task is a PDR algorithm that comprises a step detector, a stride length estimator, a heading estimator, and a dynamic gyro bias corrector. These components work in a tightly integrated manner, where the output of one component can be the input for others. For instance, the step detector's output is used by both the stride length estimator and the gyro bias monitor. Details of the algorithm will be presented in Section \ref{pdr}. 

\subsubsection{Communication Task} manages the communication between the PDRS module and the application's host controller. In this task, we implement a set of widely used communication protocols, including I2C, SPI, and UART, to ensure flexible integration of the PDRS with any system. Through these communication channels, the host controller can actively control the operation of the PDRS by sending commands or retrieving location data from its FIFO.

\section{Adapted Pedestrian Dead Reckoning } \label{pdr}
PDR aims to determine the locations of pedestrians based on an initial point \(P_0(x_0, y_0)\). To this end, when a new step $m$ is detected, it combines the corresponding stride length ( \(l_m\)) and heading (\(\psi_m\)) to produce the new location estimate \(P_m(x_m, y_m)\), where $x_m$ and $y_m$ are determined as:
\begin{equation}
    \begin{bmatrix}
        x_m\\
        y_m
    \end{bmatrix}
    =
    \begin{bmatrix}
        x_{m-1}\\
        y_{m-1}
    \end{bmatrix}
    +
    l_m \begin{bmatrix}
        \sin(\psi_m)\\
        \cos(\psi_m)
    \end{bmatrix}     \quad \quad m \geqslant 1
\end{equation}
\subsection{Stand phase detection}\label{AA}


The simplest method is peak detection \cite{6987239}, where each peak in the acceleration signal corresponds to a step. However, challenges arise in distinguishing genuine steps from other movements or noise, such as minor jitters. Factors like walking speed, terrain, and gait differences can affect the signal, leading to inaccuracies. For robust step detection, our implementation is based on state-of-the-art gait analysis methods but distinctly uses acceleration differential instead of raw acceleration. The use of acceleration differential eliminates the need for gravity removal, reducing reliance on precise device alignment and lengthy calibration, thus enhancing accuracy and practicality in gait detection. Details of the method are presented in the following steps \cite{vu2024real}:

\subsubsection{Step 1}

A $n = 4$ order low-pass filter with a cutoff frequency of $c = 2$ is used to remove the high-frequency noise of the acceleration norm. 
\begin{equation}
    a_{f, k} = f_{FIR, c, n}(\sqrt{a_{x, k}^2 + a_{y, k}^2 + a_{z, k}^2})
\end{equation}
\subsubsection{Step 2}

Ternarize the differential of $a_{f, k}$ using a pair of thresholds $\pm h (h = 0.06)$. Additionally, $w = 0.5$ removes occasional excessive wiggling that affects detection stability.
\begin{equation}
    \dot{a}_{f, k} = \frac{a_{f,k} - a_{f, k-1}}{\Delta t}
\end{equation}
\begin{equation}
    b_k= \begin{cases}
            +1 & \text{for} \quad h< \dot{a}_{f, k} <w\\ 
            -1 & \text{for} \quad \dot{a}_{f, k} < -h \quad \text{and} \quad b_{k-1}\neq 1\\ 
             0 &  \text{otherwise}
        \end{cases}
\end{equation}

\subsubsection{Step 3}
\begin{figure}[t]
  \centering
  \includegraphics[width=83mm]{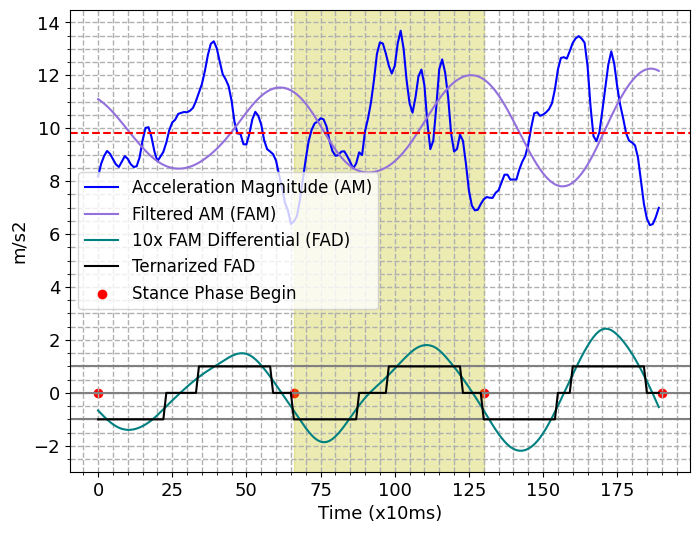}
  \caption{Detecting stance phase using acceleration differential}
  \label{fig:stance_phase_detection_using_acc_mag_differential}
\end{figure}

As shown in Figure \ref{fig:stance_phase_detection_using_acc_mag_differential}, the beginning of stance phases is determined as the moments when the $b_k$  is  -1, preceded by 0 and 1, respectively:
\begin{equation}
        s_k = 
            \begin{cases}
                1 & \text{for} \quad b_k=-1, b_{k-1}=0, b_{k-2}=1\\ 
                0 & \text{otherwise}
            \end{cases}
\end{equation}

\subsection{Stride Length Estimation}\label{BB}

Accurate stride length estimation is difficult due to the intricate nature of the musculoskeletal system, which is influenced by various factors such as age, gender, body mass, and overall fitness levels. In our study, we employed Weinberg's method \cite{weinberg2002using} that estimates stride length based on its correlation with hip bounce during walking. The accuracy of this method is  92\%  across individuals with varying leg lengths. Equation \ref{eq:stride_length} shows how the displacement is calculated. In that, \( l \) denotes the total walking distance, \( n \) is the number of steps taken, and \( a_{z, \text{min}}^{(m)} \) and \( a_{z, \text{max}}^{(m)} \) indicate the minimum and maximum accelerations observed on the Z-axis within a single stride \( m \). When estimating displacement for each individual stride, \( n \) is set to 1. Additionally, \( \pmb{K} \) (calibration is required to determine) serves as the conversion coefficient, allowing the final output to be presented in units such as feet or meters. In our implementation, the default value of \( \pmb{K} \) is 0.23 for the output in meters.
\begin{equation}
    l = \sqrt[4]{a_{z, \text{max}}^{(m)} - a_{z, \text{min}}^{(m)}} \cdot n \cdot \pmb{K}
    \label{eq:stride_length}
\end{equation}

\subsection{Heading Calculation}
\subsubsection{Heading Estimation}
The heading can be simply determined by integrating the angular velocity signals captured by the gyroscope. However, using only the gyroscope leads to accumulated error. To mitigate this error, data from another sensor can be incorporated using a complementary filter or optimal weighting algorithms\cite{ludwig2018comparison}. In this study, we employ Madgwick's method\cite{madgwick2010efficient} to calculate the orientation using data from the gyroscope and accelerometer.

The orientation is represented by a unit quaternion, $\mathbf{q}_t = [q_w,\,q_x,\,q_y,\,q_z]$, updated over time using both the angular rate and the gravitational direction. First, express the gyroscope measurements as the pure quaternion $^S\bm{\omega}_t = \begin{bmatrix} 0 & \omega_x & \omega_y & \omega_z \end{bmatrix}$, and integrate them using 
\begin{equation}
    \dot{\mathbf{q}}_{\omega,t} = \frac{1}{2}\,\mathbf{q}_{t-1}\,^S\bm{\omega}_t
\end{equation}
yielding the preliminary orientation update
\begin{equation}
    \mathbf{q}_{\omega,t} = \mathbf{q}_{t-1} + \dot{\mathbf{q}}_{\omega,t}\Delta t
\end{equation}
Here, accelerometer data is incorporated to mitigate drift. Assuming that the expected gravity vector in the Earth frame is $^E\mathbf{g} = \begin{bmatrix} 0 & 0 & 0 & 1 \end{bmatrix}$ and that the accelerometer measurement is normalized as $^S\mathbf{a} = \begin{bmatrix} 0 & a_x & a_y & a_z \end{bmatrix}$, an objective function can be defined as 
\begin{equation}
    f_g(\mathbf{q},\,^S\mathbf{a}) = 
    \begin{bmatrix}
    2(q_xq_z - q_wq_y) - a_x \\
    2(q_wq_x + q_yq_z) - a_y \\
    2\left(\frac{1}{2} - q_x^2 - q_y^2\right) - a_z
    \end{bmatrix}
\end{equation}
Its gradient, computed via the Jacobian $J_g(\mathbf{q})$, is given by
\begin{equation}
    \nabla f = J_g^T(\mathbf{q}_{t-1})\, f_g(\mathbf{q}_{t-1},\,^S\mathbf{a}_t)
\end{equation}
Orientation is updated by fusing the gyroscope integration with gradient descent correction scaled by a gain factor $\beta$:
\begin{equation}
    \mathbf{q}_t = \mathbf{q}_{t-1} + \left(\dot{\mathbf{q}}_{\omega,t} - \beta\,\frac{\nabla f}{\|\nabla f\|}\right)\Delta t
\end{equation}

Euler angles including roll ($\phi$), pitch ($\theta$), and yaw ($\psi$) are obtained from quaternion $\mathbf{q}_t$ as
\begin{equation}
\begin{aligned}
    \phi    &= \arctan2\Bigl(2(q_wq_x + q_yq_z),\,1-2(q_x^2+q_y^2)\Bigr) \\
    \theta  &= \arcsin\Bigl(2(q_wq_y - q_zq_x)\Bigr)                    \\
    \psi    &= \arctan2\Bigl(2(q_wq_z + q_xq_y),\,1-2(q_y^2+q_z^2)\Bigr) 
\end{aligned}
\end{equation}

\subsubsection{Dynamic Gyroscope Bias Update}

MEMS gyroscopes suffer bias instabilities and noise, causing drift, especially in the yaw axis. To mitigate drift, we implement a compensation mechanism that detects stationary states and updates the bias dynamically \cite{vu2024real}. If the acceleration differential lies within $\pm \varepsilon (m/s^2)$ in $T_{\text{idle}}$ (with $s$ samples).
\begin{equation}
   -\varepsilon \leq a_{f, i+1} - a_{f, i} \leq \varepsilon \quad \forall  i \in [k-s+1, k]
\end{equation}
With $s$ samples of angular rate $\bm{\omega}_i = \begin{bmatrix} \omega_x & \omega_y & \omega_z \end{bmatrix}$ collected in sensor frame during the stationary interval $ G_{\text{idle}} = \{\bm{\omega}_{k-s+1},..., \bm{\omega}_{k}\}$, the initial gyroscope bias $\mathbf{b}_0 = [b_x \quad b_y \quad b_z ]^T$ will be calculated when the first $s_{\text{smin}}$ (where $s_{\text{smin}} < s$) samples is detected.
\begin{gather}
    \mathbf{b}_0 = \frac{\sum_{i=k-s +1}^{k-s+s_{\text{min}}}\bm{\omega}_i}{s_{\text{min}}}
\end{gather}
Gyroscope bias is recalculated every $w$ (adjustable) samples ($1 \le w < s_{\text{min}}$), with the update window extending until the idle state ends. This prevents abnormal bias from sharp gyro‐signal spikes during idle‐to‐motion transitions while preserving the update frequency compared to using a long fixed window.

\begin{equation}\label{eq:calculate_bias_using_ext_window}
\begin{aligned}
\resizebox{1.\hsize}{!}{$
    \mathbf{b}_{m} = \frac{(s_{\text{min}} + m\times w -w) \times \mathbf{b}_{m-1} + \sum_{k-s+s_{\text{min}}+(m-1)\times w+1}^{k-s+s_{\text{min}}+m\times w}\bm{\omega}_i}{s_{\text{min}} + m\times w} 
    $} \\
    m\in[1, 2,...]
\end{aligned}
\end{equation}
The corrected angular rate \(\bm{\omega}_{\text{deb}, i} = \bm{\omega}_{i} - \mathbf{b}_m\), applicable when \(i > k-s+s_{\text{min}}+m \times w\), can be utilized to improve the heading estimation.

\section{Experiments and Results} \label{experiments_and_results}

\subsection{Experimental Setup}

\begin{figure}[t]
  \centering
  \includegraphics[width=80mm]{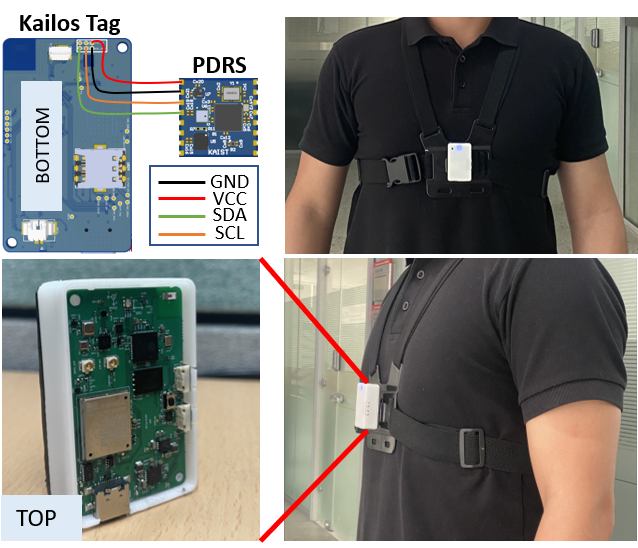}
  \caption{Integration of the PDRS on a wearable device}
  \label{fig:experimental_setup}
\end{figure}

To evaluate the PDRS, we leverage a wearable device developed for our KAILOS positioning system \cite{kailos, vu2022iot, vu2024real}. This device, powered by a dual-core 32-bit processor, includes BLE, WiFi, LTE, GNSS, and various sensors. The PDRS module connects to the device’s main board through a four-wire interface: two wires (VCC and GND) supply power, while the other two (SDA and SCL) enable I2C communication. Figure \ref{fig:experimental_setup} shows the device worn on the body and the connection circuit between the device and the PDRS.

\subsection{Indoor and outdoor positioning result}

\begin{table}[b]
\renewcommand{\arraystretch}{1.3}
\caption{Comparison of positioning results between PDRS and several systems and devices}
\label{tab:compare_pdrs}
\resizebox{\columnwidth}{!}{%
\begin{tabular}{|l|c|c|c|c|c|}
\hline
                                                                          & \textbf{\begin{tabular}[c]{@{}c@{}}Missed \\ Steps\\ (over 1250\\ Step)\end{tabular}} & \textbf{\begin{tabular}[c]{@{}c@{}}SDE(\%)\\ (over 1250 \\ steps)\end{tabular}} & \textbf{\begin{tabular}[c]{@{}c@{}} Max \\ TDE (\%) \\ (over 34 \\ trials)\end{tabular}} & \textbf{\begin{tabular}[c]{@{}c@{}}Max \\ LCE (m) \\ (over 34 \\trials)\end{tabular}} & \textbf{Note}                                                                   \\ \hline
\textbf{PDRS}                                                             & 2                                                                                    & 0.16                                                                          & 1.87              & 3.5               &                                                                                 \\ \hline
\textbf{\begin{tabular}[c]{@{}l@{}}Smartphone \\ PDR\end{tabular}}        & 5                                                                                    & 0.40                                                                          & 6.15              & 13.6              &                                                                                 \\ \hline
\textbf{\begin{tabular}[c]{@{}l@{}}Openshoes \\ (ZUPT)\end{tabular}}      & 2                                                                                    & 0.16                                                                          & 5.21              & 7.1               & \begin{tabular}[c]{@{}c@{}}Only algorithm \\ is used\end{tabular}              \\ \hline
\textbf{\begin{tabular}[c]{@{}l@{}}BHI160B/\\ BHI260AP\end{tabular}}      & 52                                                                                   & 4.16                                                                          & NA                & NA                & \begin{tabular}[c]{@{}c@{}}PDR-Enable \\ Firmware\\ is unavailable \cite{BHY2_SensorAPI}\end{tabular} \\ \hline
\textbf{LSM6DS3}                                                          & 55                                                                                   & 4.4                                                                           & NA                & NA                & ST                                                                              \\ \hline
\textbf{BHI360}                                                           & 54                                                                                   & 4.32                                                                          & NA                & NA                & Bosch                                                                           \\ \hline
\textbf{\begin{tabular}[c]{@{}l@{}}Galaxy  Note \\ 20 Ultra\end{tabular}} & 2                                                                                    & 0.16                                                                          & NA                & NA                & Samsung                                                                         \\ \hline
\textbf{\begin{tabular}[c]{@{}l@{}}Apple Watch \\ Ultra\end{tabular}}     & 20                                                                                   & 1.6                                                                           & NA                & NA                & Apple                                                                           \\ \hline
\textbf{\begin{tabular}[c]{@{}l@{}}Galaxy\\ Watch 7\end{tabular}}         & 28                                                                                   & 2.24                                                                          & NA                & NA                & Samsung                                                                         \\ \hline
\end{tabular}%
}
\end{table}        

PDRS's positioning accuracy was evaluated over 26 trials on the KAIST campus (Daejeon, Korea) in both indoor (nine‑storey N1 building; 114.6 m × 39.6 m) and outdoor (where GNSS signal is well received)  settings. Closed-loop trajectories of 135–655 m were traversed by different individuals at unscripted gait speeds (slow, normal, and fast) in both clockwise and counter‑clockwise directions to eliminate directional bias. The PDRS was benchmarked against academic prototypes, commercial off‑the‑shelf units, and consumer‑grade devices using three main metrics: Step‑Detection Error (SDE), Travelled‑Distance Error (TDE), and loop‑closure error (LCE).

As shown in Table~\ref{tab:compare_pdrs}, with only two missed steps in a 1250‑step trajectory \(\bigl(0.16\%\ \text{SDE}\bigr)\), PDRS matches the foot‑mounted OpenShoe ZUPT system in step detection yet yields a markedly smaller distance error \(\bigl(1.87\% \text{ vs.\ } 5.21\%\bigr)\) and a tighter loop‑closure error \(\bigl(3.5\,\text{m vs.\ } 7.1\,\text{m}\bigr)\). Compared with a conventional smartphone‑based PDR implementation, PDRS cuts SDE by \(60\%\), reduces TDE by a factor of \(3.3\), and shrinks LCE from \(13.6\,\text{m}\) to \(3.5\,\text{m}\), underscoring the benefits of its acceleration‑differential step detector and dynamic gyro‑bias compensation. Notably, we observed that both the smartphone PDR and the OpenShoe ZUPT algorithm suffer pronounced accuracy degradation outdoors, where uneven terrain and larger operating areas amplify drift; this environment‑specific weakness largely explains the wide performance gap between those systems and PDRS. Commodity IMU chips such as Bosch’s BHI160B/BHI260AP (used without PDR‑enabled firmware) and ST’s LSM6DS3 miss more than fifty steps \(\bigl(\approx 4\%\,\text{SDE}\bigr)\) and cannot provide reliable displacement estimates, while recent flagship smartphones and smartwatches exhibit wide variability \(\bigl(0.16\text{–}2.24\%\,\text{SDE}\bigr)\) and likewise lack unified trajectory outputs. Overall, the results indicate that wearable integration of PDRS yields reliable, precise positioning regardless of the environment.

\begin{figure}[t]
  \centering
  \includegraphics[width=85mm]{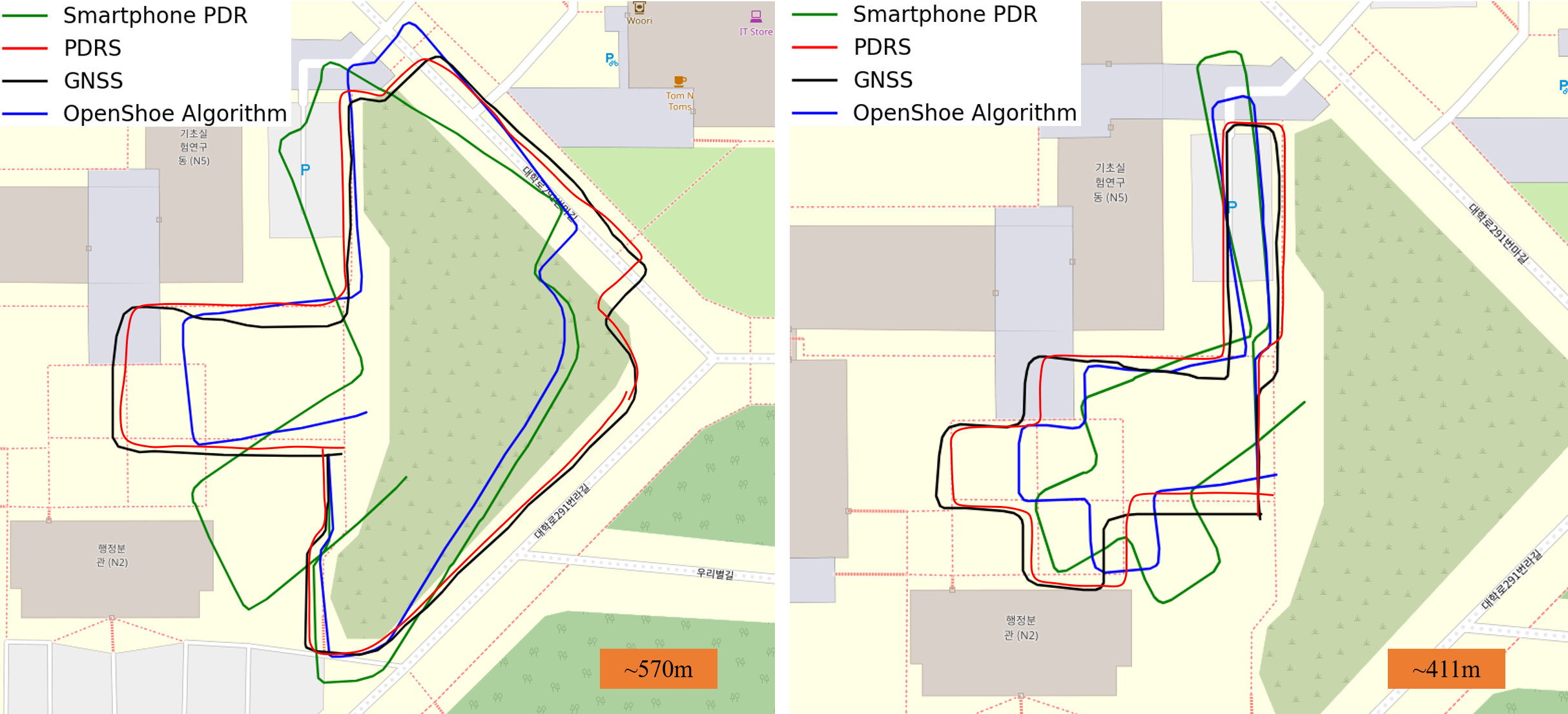}
  \caption{Outdoor positioning results of the PDRS}
  \label{fig:pdr_eval_out}
\end{figure}

\begin{figure}[t]
  \centering
  \includegraphics[width=85mm]{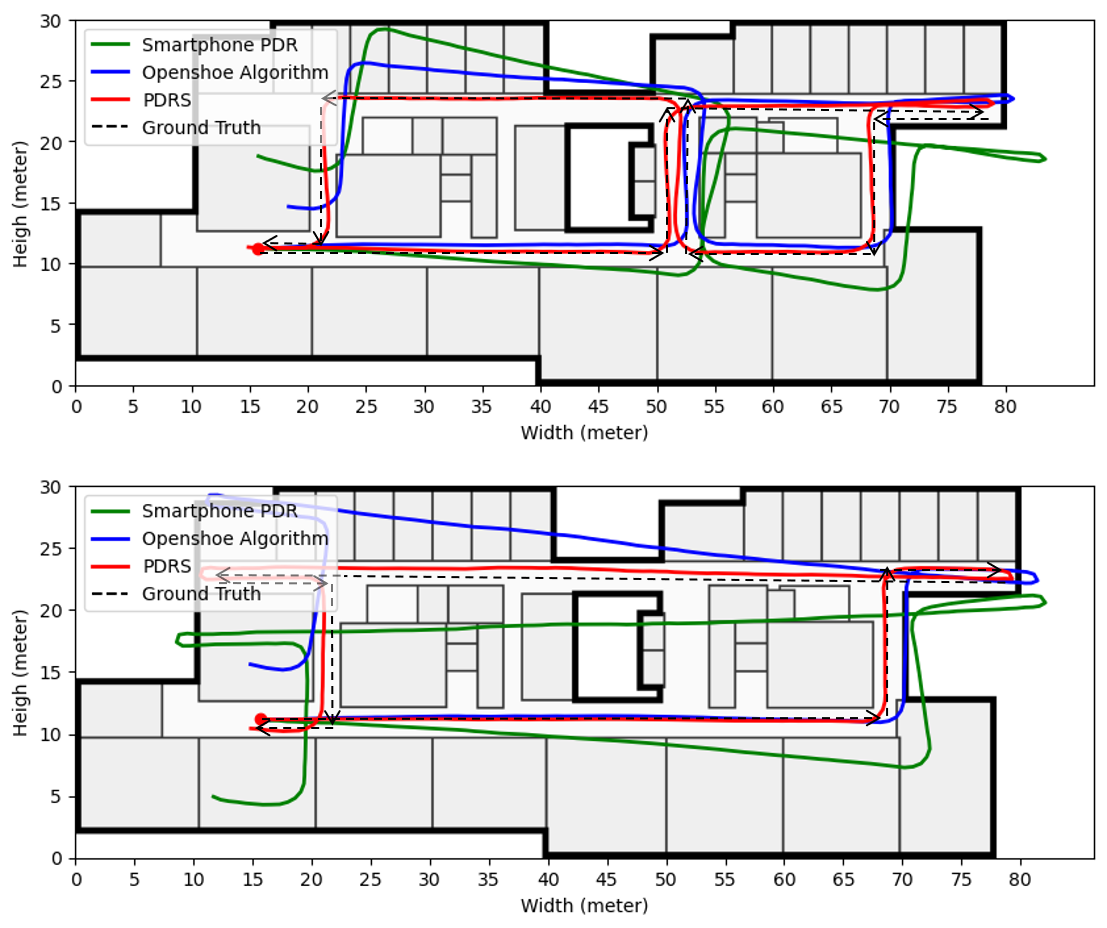}
  \caption{Indoor positioning results of the PDRS}
  \label{fig:pdr_track}
\end{figure}

Figures\ref{fig:pdr_track} and \ref{fig:pdr_eval_out} depict several estimated trajectories of both indoor and outdoor trials. Indoor ground‑truth paths were digitised from floor plans, whereas outdoor references were captured with real‑time‑kinematic (RTK) GPS. In every scenario, the PDRS outputs align markedly closer to the ground truth than the trajectories generated by other systems.

\subsection{Operation Evaluation}



\subsubsection{Computational Cost}
\begin{figure}[t]
  \centering
  \includegraphics[width=80mm]{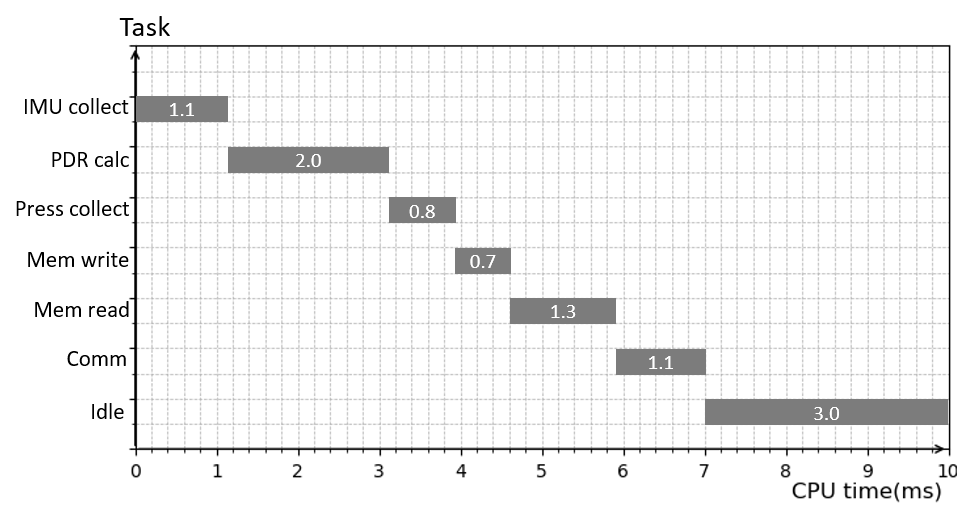}
  \caption{CPU usage of the tasks within 10 milliseconds}
  \label{fig:task_execution_tiome}
\end{figure} 
Figure \ref{fig:task_execution_tiome} presents the CPU utilization of individual tasks, profiled over a $10$ ms interval. The vertical axis lists tasks in order of increasing priority level, while the horizontal axis indicates the execution time allocated to each task. Excluding the idle task, the PDR computation requires the most CPU time, consuming $2.0$ ms (approximately $20\%$ of the interval). The task with the lowest usage is the write operation, at $0.7$ ms. The high-priority IMU sampling and communication tasks each occupy $1.1$ ms. Meanwhile, pressure sensing and memory read tasks consume $0.8$ ms and $1.3$ ms, respectively. These results demonstrate that the PDR algorithm imposes a moderate computational load relative to other system functions, confirming that the PDRS can maintain stable real-time performance within the available processing budget.
\begin{figure}[b]
  \centering
  \includegraphics[width=80mm]{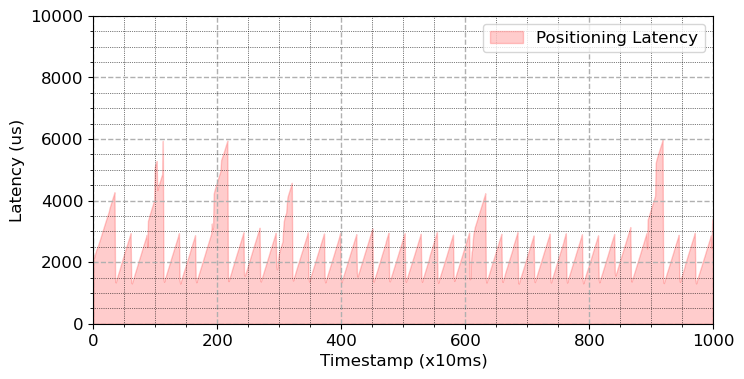}
  \caption{Positioning latency of the PDRS}
  \label{fig:pdr_latency}
\end{figure}

\subsubsection{Positioning Latency}

Positioning latency is defined as the interval between the moment raw sensor data are captured and the moment the corresponding position estimate is computed and becomes available. This latency encompasses various stages, including data acquisition, preprocessing, execution of the PDR algorithm, and buffering or transmission overhead. Through firmware-level optimizations, such as interrupt-driven sampling, fixed-point arithmetic, and pipelined processing, the end-to-end latency has been significantly reduced to an average of $2.5$ ms. Figure~\ref{fig:pdr_latency} presents a latency trace over a representative interval of $10$ s, illustrating a worst-case delay of $6$ ms. Over an extended ten-hour continuous testing period, only a few instances exhibited latency exceeding $8$ ms, with none surpassing $10$ ms. Given typical pedestrian walking frequencies of $2$–$3$ steps per second, the positioning latency achieved by the PDRS is sufficiently low to be considered negligible.

\subsubsection{Power Consumption}
\begin{figure}[t]
  \centering
  \includegraphics[width=80mm]{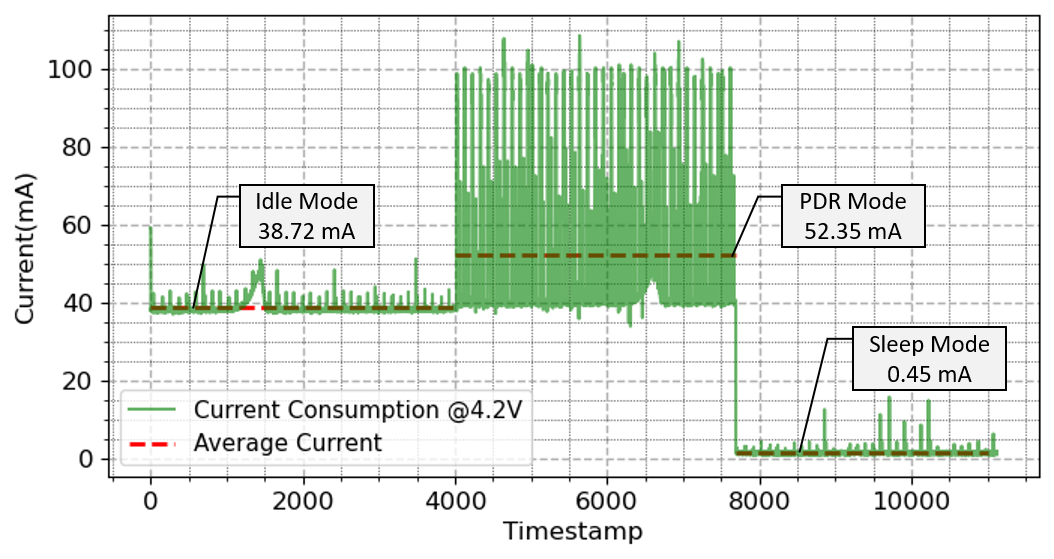}
  \caption{Power consumption of the PDRS module in different modes}
  \label{fig:power_consuption}
\end{figure}

Power consumption is a primary design constraint when embedding the PDRS in battery‑powered platforms. In this evaluation, we quantified the supply current at a regulated $4.2\,\text{V}$ across three operating modes; the results are summarized in Figure~\ref{fig:power_consuption}. Specifically, in \emph{idle} mode, with all sensors powered but computation suspended, the module draws $38.72\,\text{mA}$. Activating \emph{PDR} mode that continuously collect sensor data and estimate position, increases the load by only $13.63\,\text{mA}$, resulting in a total of $52.35,\text{mA}$. Conversely, entering \emph{sleep} mode reduces the current to just $0.45\,\text{mA}$. This profile is adequate for most wearable and handheld applications.

\subsubsection{Thermal Performance}

\begin{figure}[b]
  \centering
  \includegraphics[width=80mm]{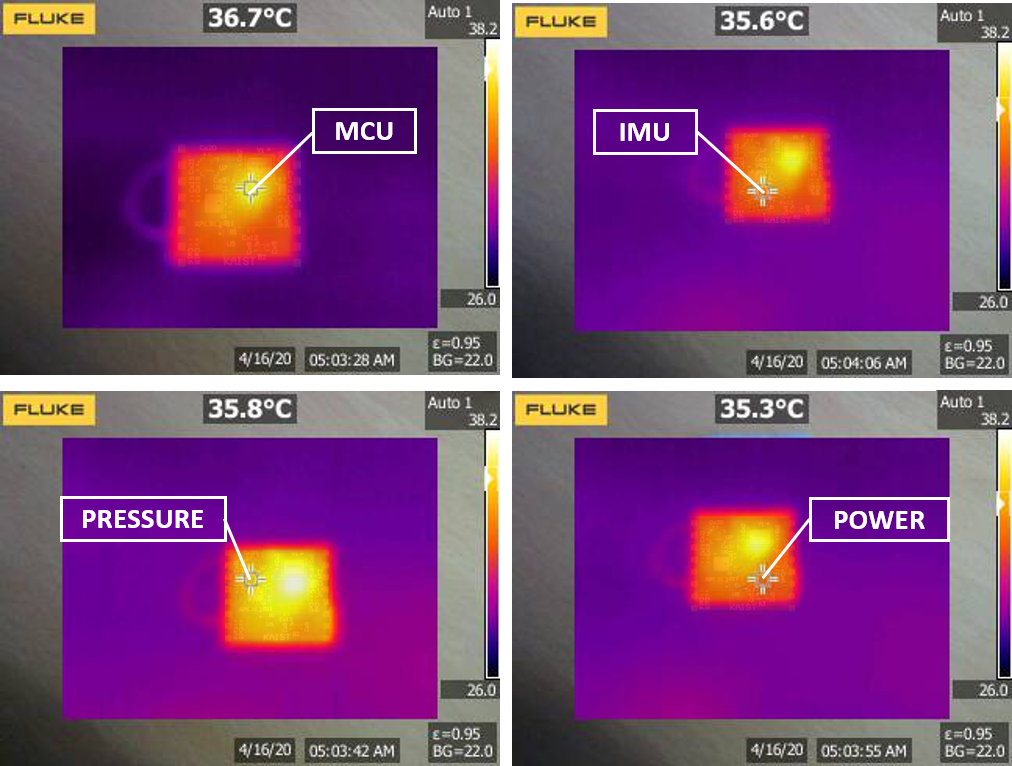}
  \caption{Temperature distribution across sensors and components of PDRS}
  \label{fig:heat_model}
\end{figure}

For reliable operation in wearable platforms, the thermal behavior of the PDRS module was evaluated under steady-state conditions using a Fluke TiS60+ thermal imager. Figure~\ref{fig:heat_model} presents the resulting temperature distribution. In that, the microcontroller (MCU) exhibited the highest temperature at $36.7^{\circ}\text{C}$, attributable to its continuous execution of PDR algorithms. The IMU, sampling at $100\text{Hz}$, stabilized at $35.6^{\circ}\text{C}$, while the barometric pressure sensor reached $35.8^{\circ}\text{C}$—likely a consequence of its metal package and proximity to the MCU. The low-dropout voltage regulator remained the coolest component at $35.3^{\circ}\text{C}$, confirming its efficient power conversion.

These measurements demonstrate effective heat dissipation, with all components operating within the thermal limits required for wearable electronics \cite{IEC62368-1}. In particular, maintaining the IMU below critical temperature thresholds is essential, as thermal drift can introduce measurement bias and degrade accuracy \cite{patrizi2024temperature}. The observed thermal stability ensures that the PDRS can deliver precise position estimates without requiring additional cooling or performance derating.
\section{Conclusion and Discussion} \label{conclusion_and_discussion}

This paper presented a black-box PDR solution that seamlessly integrates a real-time PDR algorithm into a compact, low-power, and cost-effective module. By abstracting the complexities of sensor fusion and computation logic within a plug-and-play design, the proposed PDRS significantly simplifies the integration of robust indoor localization capabilities into wearable and IoT systems. Extensive evaluations in both indoor and outdoor settings demonstrated competitive performance with a step detection error of 0.16\%, a travel distance error of 1.87\%, and a loop-closure error of 3.5 m  while maintaining average positioning latency of 2.5 ms and an efficient power profile.

Future research will pursue three core objectives to advance the PDRS. First, we aim to optimize both hardware and software to enhance positioning accuracy while minimizing power consumption. Second, we plan to adopt ASIC technology to shrink the PDRS to a chip-scale form factor, reducing production costs and enabling integration into smaller devices. Third, building on the success of deep learning in localization, we will develop an AI-enhanced PDRS incorporating a dedicated model tailored to our hardware. This AI-driven approach will enable the sensor to better compensate for drift by learning sensor dynamics from extensive datasets, thereby delivering superior real-time accuracy even in complex environments.

\bibliographystyle{IEEEtran}
\bibliography{reference}

\end{document}